%% file: main.tex
\documentclass[11pt]{article}

\usepackage[utf8]{inputenc}
\usepackage[T1]{fontenc}
\usepackage{lmodern}
\usepackage{microtype}
\usepackage[a4paper,margin=2.5cm]{geometry}
\usepackage{amsmath,amssymb,amsthm,mathtools}
\usepackage{bm}
\usepackage{graphicx}
\usepackage{booktabs}
\usepackage{multirow}
\usepackage{array}
\usepackage{xcolor}
\usepackage[ruled,vlined,linesnumbered]{algorithm2e}
\usepackage[numbers,sort&compress]{natbib}
\usepackage[colorlinks=true,linkcolor=blue,citecolor=blue,urlcolor=blue]{hyperref}
\usepackage{orcidlink}
\usepackage{cleveref}
\usepackage{placeins}

\graphicspath{{figures/}}


\newtheorem{theorem}{Theorem}
\newtheorem{proposition}[theorem]{Proposition}
\newtheorem{lemma}[theorem]{Lemma}
\newtheorem{corollary}[theorem]{Corollary}
\theoremstyle{definition}
\newtheorem{definition}{Definition}

\theoremstyle{remark}
\newtheorem{remark}{Remark}

\newcommand{\R}{\mathbb{R}}

\newcommand{\Pb}{\mathbb{P}}
\newcommand{\D}{\mathcal{D}}

\newcommand{\ind}{\mathbf{1}}
\newcommand{\eank}{\textup{\textsc{EANK}}}
\DeclareMathOperator*{\argmax}{arg\,max}

\DeclareMathOperator{\rnd}{rnd}
\newcommand{\doi}[1]{\href{https://doi.org/#1}{doi:\,#1}}

\title{Evolutionary Adaptive Neighborhood Learning: A Genetic Algorithm for Instance-Dependent $k$ Optimization in Nearest Neighbor Classification}

\author{%
  Isaac Touza\textsuperscript{1,2,$*$}\,\orcidlink{0009-0003-6288-9458},\quad
  Kaladzavi Guidedi\textsuperscript{2,3}\,\orcidlink{0000-0002-8414-214X},\quad
  Kolyang\textsuperscript{2,4}\,\orcidlink{0000-0003-1554-1187}\\[2ex]
  \parbox{0.92\linewidth}{\centering\small
    \textsuperscript{1}Department of Mathematics and Computer Science, Faculty of Science, University of Maroua, P.O.~Box 814, Maroua, Cameroon\\[0.4ex]
    \textsuperscript{2}Laboratoire de Recherche en Informatique (LaRI), University of Maroua, Maroua, Cameroon\\[0.4ex]
    \textsuperscript{3}Department of Computer Science and Telecommunications, National Advanced School of Engineering, University of Maroua, P.O.~Box 46, Maroua, Cameroon\\[0.4ex]
    \textsuperscript{4}Higher Teacher's Training College, University of Maroua, P.O.~Box 46, Maroua, Cameroon}%
}
\date{}

\begin{document}
\maketitle
{\renewcommand{\thefootnote}{\fnsymbol{footnote}}%
\footnotetext[1]{Corresponding author. E-mail: \href{mailto:isaac_touza@outlook.fr}{isaac\_touza@outlook.fr}.}}

\begin{abstract}
The number of neighbours $k$ is the central hyper-parameter of the $k$-nearest
neighbour ($k$-NN) classifier, and it is almost always chosen \emph{globally}, by
cross-validation or, increasingly, by evolutionary search. Yet the bias--variance
trade-off that governs $k$ is local: dense, noise-free regions favour small
neighbourhoods, whereas sparse or label-noisy regions favour large ones. Because
the global-$k$ search space is one-dimensional, replacing exhaustive search by a
genetic algorithm (GA) cannot improve on it. We therefore change the object being
optimised. We propose \eank{} (Evolutionary Adaptive Neighborhood $K$), which
evolves a low-dimensional \emph{policy} $g_\theta$ mapping a description of the
local neighbourhood of a point (density, label entropy at two scales, vote
margin, concentration, relative density) to an instance-dependent number of
neighbours $k(x)\in\{1,\dots,K\}$. A real-coded elitist GA maximises the
leave-one-out accuracy of the induced adaptive rule, penalised by a
neighbourhood-smoothness term. A pre-computed vote table reduces each fitness
evaluation to $O(n)$ look-ups. We prove that (i)~a seeded \eank{} never does worse
than the best global $k$ on its training criterion, (ii)~the GA converges almost
surely to the essential supremum of the fitness, (iii)~the learned neighbourhood
size is Lipschitz in the meta-features, and (iv)~the policy class has a
growth function polynomial in the sample size with exponent $d+2$. This yields a
uniform generalisation bound that does not depend on $n$, which per-instance
encodings do not enjoy. On a controlled benchmark with spatially heterogeneous
structure, \eank{} recovers small neighbourhoods in the fine-grained region and
large ones in the noisy region, and it improves accuracy over the best global
$k$ at every noise level. In a nested, $5\times5$-fold evaluation on twelve
public datasets, \eank{} obtains the best average rank among six $k$-NN variants.
Its accuracy is statistically indistinguishable from the tuned global-$k$
baselines, and it is significantly better than the default $k=5$. We document
the gap between in-sample and out-of-sample gains, and we identify it as the
main obstacle for adaptive-neighbourhood methods.
\end{abstract}


\section{Introduction}

The $k$-nearest neighbour rule \citep{fix1951,silverman1989,cover1967} remains one of the most
widely used non-parametric classifiers. It is simple and universally consistent
when $k\to\infty$ and $k/n\to0$ \citep{stone1977,devroye1996}, and its
predictions are easy to interpret. Its behaviour, however, depends strongly
on the neighbourhood size $k$. A small $k$ gives a low-bias, high-variance rule
that is sensitive to label noise. A large $k$ smooths the decision boundary but
erases fine structure and minority regions.

In practice $k$ is selected \emph{once for the whole input space}, by grid search
with cross-validation, by leave-one-out (LOO) error, or by a metaheuristic. A
recent line of applied work uses genetic algorithms (GAs) to tune $k$, often
jointly with feature selection or distance weights
\citep{kelly1991,raymer2000,prasetio2020,mcroberts2016}. Further studies examine
how the choice of $k$ interacts with feature selection and class imbalance
\citep{dawkins2025}. When $k$ is the only quantity being optimised, the search
space $\{1,\dots,K\}$ is one-dimensional and small. Exhaustive LOO search costs
essentially the same as a single $k$-NN pass, because all $K$ predictions of a
point can be read from one sorted neighbour list. Exhaustive search therefore
finds the exact optimum of the global-$k$ criterion, and it is an upper bound
for any other optimiser of that criterion: a GA, random search
\citep{bergstra2012} or Bayesian optimisation. Replacing grid search by a GA
cannot, by construction, improve on it.

The statistical theory of nearest neighbours, by contrast, says that the right
amount of smoothing is \emph{local}. Convergence rates depend on the local
density and the local margin \citep{chaudhuri2014}, $k$-NN adapts to the local
intrinsic dimension \citep{kpotufe2011}, and recent adaptive rules choose the
neighbourhood point by point \citep{balsubramani2019}. This suggests changing
the object being optimised: we look for a \emph{function} $x\mapsto k(x)$ rather
than an integer.

\paragraph{Approach.} We parameterise this function as a \emph{policy}
$k_\theta(x)=1+\rnd\!\big((K-1)\,\sigma(w^\top z(x)+b)\big)$. Here $z(x)\in\R^{d}$
is a small vector of \emph{meta-features} that describe the neighbourhood of $x$
in the training set (scale, label entropy, vote margin, concentration, relative
density), and $\theta=(w,b)\in\R^{d+1}$ with $d=7$. The induced rule classifies
$x$ by majority vote among its $k_\theta(x)$ nearest neighbours. The objective is
piecewise constant and non-differentiable in $\theta$, and it has many plateaus,
so we optimise it with a real-coded elitist GA. The fitness is the LOO accuracy
of the adaptive rule minus a smoothness penalty on $k$ between neighbouring
points.

\paragraph{Contributions.}
\begin{enumerate}
\item \textbf{Formulation.} We recast GA-based tuning of $k$ as learning an
  instance-dependent neighbourhood policy. The policy has a fixed, small number
  of parameters, independent of $n$, and it contains every global-$k$ rule as a
  special case (\Cref{sec:method}).
\item \textbf{Algorithm.} \eank{} combines interpretable meta-features, a smooth
  monotone policy and an elitist GA with simulated binary crossover
  \citep{deb1995,deb2001}, annealed Gaussian mutation and seeding with constant policies.
  A pre-computed LOO vote table makes a fitness evaluation cost $O(n)$, so a full
  evolution takes about a tenth of a second on the datasets considered
  (\Cref{sec:method,sec:complexity}).
\item \textbf{Theory.} We prove that the seeded GA dominates the best global $k$
  on its training criterion, that the GA converges almost surely, and that the
  learned neighbourhood size is stable. Through a Sauer--Shelah argument we also
  obtain a uniform generalisation bound for the policy class whose complexity
  term depends on $d$ and $\log K$, not on $n$. The bound separates the policy
  approach from per-instance encodings (\Cref{sec:theory}).
\item \textbf{Evidence.} We report a controlled synthetic study and a nested
  $5\times5$-fold benchmark on twelve public datasets against five $k$-NN
  baselines, including a classical GA over $k$ and a two-stage local-$k$ method.
  We add ablations and corrected statistical tests (\Cref{sec:experiments}).
  We report negative and null results alongside positive ones.
\end{enumerate}

\section{Related work}\label{sec:related}

\paragraph{Choosing a global $k$.} Classical recommendations range from
$k\approx\sqrt n$ to LOO or $v$-fold cross-validation. The LOO error of $k$-NN
admits distribution-free performance bounds \citep{devroye1979,devroye1996}, and
\citet{ghosh2006} studied the optimal choice of $k$ from a Bayesian viewpoint.
Distance weighting \citep{dudani1976} and optimally weighted schemes
\citep{samworth2012} soften the dependence on $k$ without removing it.

\paragraph{Locally adaptive neighbourhoods.} \citet{wettschereck1994} proposed
locally adaptive nearest neighbour algorithms that store, for each training
point, the values of $k$ that classify it correctly and reuse them for nearby
queries. \citet{hastie1996} adapted the \emph{metric} locally (DANN).
\citet{wang2006} chose the neighbourhood size from a statistical confidence
criterion. \citet{zhang2018ktree} learned an optimal $k$ for every training
sample by sparse reconstruction and then fitted a decision tree (kTree) to
predict $k$ for test points. \citet{mullick2018} learned point-specific $k$ with
a neural network, with a class-dependent weighting for imbalanced data. On the
theory side, \citet{balsubramani2019} gave an adaptive rule with point-wise
guarantees that grows the neighbourhood until a statistical bias test is
passed, and \citet{chaudhuri2014,kpotufe2011} showed that the optimal smoothing
is governed by local quantities. These methods either learn targets point by
point, which risks overfitting the LOO labels, or use a hand-designed stopping
test. \eank{} sits between the two: a hand-designed, interpretable
\emph{representation} of the neighbourhood, and a \emph{learned} map from that
representation to $k$, fitted directly on the end-to-end adaptive LOO accuracy.

\paragraph{Evolutionary optimisation of $k$-NN.} GAs were combined with $k$-NN
early on to learn feature weights \citep{kelly1991}, and later for joint feature
selection and extraction \citep{raymer2000}. More recent applied studies use GAs
to tune $k$ and select features on medical data \citep{prasetio2020}, or to
optimise variables, metrics and weighting schemes in forest inventory
\citep{mcroberts2016}. To our knowledge, these approaches all treat $k$ as a
single global gene. The genome of \eank{} instead encodes a decision rule
\emph{for} $k$. For the theory we rely on the classical convergence analysis of
elitist GAs \citep{rudolph1994} and on Vapnik--Chervonenkis theory
\citep{vapnik1971,sauer1972,devroye1996}.

\section{Problem formulation}\label{sec:problem}

Let $\D=\{(x_i,y_i)\}_{i=1}^n$ be an i.i.d.\ sample from a distribution $\Pb$ on
$\R^p\times\{1,\dots,C\}$. For a query $x$, let $x_{(1)}(x),x_{(2)}(x),\dots$ be the
training points sorted by increasing Euclidean distance to $x$. For training
points, the query itself is excluded (LOO). Let $y_{(r)}(x)$ denote their
labels. For $k\le K$ the $k$-NN prediction is
\begin{equation}
  \hat y_k(x)=\argmax_{c}\; \sum_{r=1}^{k}\ind[y_{(r)}(x)=c],
  \label{eq:knn}
\end{equation}
with ties broken towards the class of the closer neighbours (\Cref{sec:votetable}).

\begin{definition}[Adaptive neighbourhood rule]
A \emph{neighbourhood policy} is a map $g:\R^p\to\{1,\dots,K\}$. The
associated rule is $h_g(x)=\hat y_{g(x)}(x)$. Global $k$-NN is the special case
$g\equiv k$.
\end{definition}

The ideal policy minimises the risk $R(g)=\Pb\{h_g(X)\neq Y\}$. We learn $g$ by
minimising a regularised empirical criterion
\begin{equation}
  \mathcal J(\theta)=\underbrace{\frac1n\sum_{i=1}^n \ind\big[\hat y^{(-i)}_{k_\theta(x_i)}(x_i)\neq y_i\big]}_{\text{LOO error of the adaptive rule}}
  \;+\;\lambda\,\Omega(\theta),\qquad
  \Omega(\theta)=\frac{1}{nq}\sum_{i=1}^n\sum_{j\in N_q(i)}\frac{|k_\theta(x_i)-k_\theta(x_j)|}{K-1},
  \label{eq:objective}
\end{equation}
where $\hat y^{(-i)}$ uses $\D\setminus\{(x_i,y_i)\}$ and $N_q(i)$ is the set of
$q$ nearest training neighbours of $x_i$. The penalty $\Omega\in[0,1]$
discourages abrupt changes of neighbourhood size between nearby points. Without
it, the optimiser could fit idiosyncratic LOO outcomes. Equivalently, we
\emph{maximise} the fitness $F(\theta)=1-\mathcal J(\theta)$.

Two parameterisations of $g$ are possible. A \emph{direct} encoding stores one
$k_i$ per training point and needs a separate rule to extend it to new points,
as in \citet{wettschereck1994,zhang2018ktree}. Its dimension grows with $n$ and,
as discussed in \Cref{sec:theory}, it has no sample-size-free capacity control.
\eank{} uses a \emph{policy} encoding with $d+1$ parameters, introduced next.

\section{The EANK method}\label{sec:method}

\subsection{Neighbourhood meta-features}

For every point $x$ (training points in LOO mode), we compute from its sorted
neighbour list the following $d=7$ meta-features, with a local scale $m$ (default
$10$) and a wider scale $M$ (default $30$). Here $\delta_r(x)$ is the distance to
the $r$-th neighbour and $n_c^{(m)}(x)$ is the number of neighbours of class $c$
among the first $m$.
\begin{enumerate}
\item \texttt{log\_dm}: $\log\delta_m(x)$, an inverse local density.
\item \texttt{entropy\_m}: $H\big(n^{(m)}(x)/m\big)/\log C$, the normalised label entropy among $m$ neighbours.
\item \texttt{margin\_m}: $\big(n^{(m)}_{[1]}(x)-n^{(m)}_{[2]}(x)\big)/m$, the vote margin between the two leading classes.
\item \texttt{nn\_agree}: $\ind\big[y_{(1)}(x)=\argmax_c n_c^{(m)}(x)\big]$, whether the 1-NN agrees with the $m$-NN majority.
\item \texttt{concentration}: $\log\delta_1(x)-\log\delta_m(x)\le 0$, which is very negative when an almost-duplicate neighbour exists.
\item \texttt{rel\_density}: $\log\delta_m(x)-\frac1m\sum_{r\le m}\log\delta_m(x_{(r)})$, an outlierness score relative to the neighbours' own scale.
\item \texttt{entropy\_M}: the label entropy among $M$ neighbours, a second, coarser scale.
\end{enumerate}
The features are standardised with means and deviations computed on the
training fold. They are cheap by-products of the neighbour search that $k$-NN
performs anyway. None of them uses the label of the query, so test-time
meta-features are computed exactly like training-time ones.

\subsection{Policy}

Given $z(x)\in\R^d$ and $\theta=(w,b)$,
\begin{equation}
  k_\theta(x)=1+\rnd\!\Big((K-1)\,\sigma\big(w^\top z(x)+b\big)\Big),\qquad \sigma(s)=\frac{1}{1+e^{-s}},
  \label{eq:policy}
\end{equation}
where $\rnd(a)=\lfloor a+\tfrac12\rfloor$. The policy is monotone in a single
linear score, which keeps it interpretable: the sign of $w_j$ says whether
meta-feature $j$ calls for larger or smaller neighbourhoods. The constant
policies $w=0$ recover every global $k$. Parameters live in the box
$\Theta=[-B_w,B_w]^d\times[-B_b,B_b]$ ($B_w=4$, $B_b=6$).

\subsection{Vote table and fitness evaluation}\label{sec:votetable}

The cost of evaluating $F$ is dominated by computing the $n$ LOO predictions
$\hat y^{(-i)}_{k_i}(x_i)$ for arbitrary vectors $(k_i)$. We pre-compute once the
cumulative vote tensor $V_{i,k,c}=\sum_{r\le k}\ind[y_{(r)}(x_i)=c]$. A secondary
tie-breaking score $T_{i,k,c}=\sum_{r\le k}\ind[y_{(r)}(x_i)=c]/r$ favours closer
neighbours. Since $T\le H_K$ (the $K$-th harmonic number), the combined score
$V+T/(H_K+1)$ never lets $T$ override a full vote. The prediction and
correctness tables
\begin{equation}
  P_{ik}=\argmax_c\Big(V_{i,k,c}+\tfrac{T_{i,k,c}}{H_K+1}\Big),\qquad
  A_{ik}=\ind[P_{ik}=y_i]
\end{equation}
are stored in $O(nK)$ memory. Then
$F(\theta)=\frac1n\sum_i A_{i,k_\theta(x_i)}-\lambda\Omega(\theta)$ costs $O(n(d+q))$.
The global-$k$ LOO curve is the column mean $\bar A_{\cdot k}$, so the best
global $k^\star=\argmax_k\bar A_{\cdot k}$ comes for free.

\subsection{Genetic algorithm}

\Cref{alg:eank} summarises the method. We use a population of $P=40$ real-coded
genomes, binary tournament selection of size 3, simulated binary crossover
(SBX) \citep{deb1995,deb2001} with probability $0.9$ and index $\eta=15$, and Gaussian
mutation. Each gene mutates with probability $1/D$ ($D=d+1$), with at least one
gene per offspring, and the step size is
$\sigma_g=\sigma_0(\sigma_{\min}/\sigma_0)^{g/G}$ relative to the box width
($\sigma_0=0.5$, $\sigma_{\min}=0.02$). Offspring are reflected into $\Theta$.
Replacement is elitist ($2$ elites), and evolution stops after $G=60$ generations
or $25$ generations without improvement. The initial population is
\emph{seeded} with the constant policies encoding $k^\star$ and
$k\in\{1,3,5,9,15,25,K\}$. The remaining genomes are drawn uniformly in $\Theta$.

\begin{algorithm}[t]
\caption{\eank{}: Evolutionary Adaptive Neighborhood $K$}\label{alg:eank}
\KwIn{training set $\D$, $K$, scales $m,M$, smoothness $(\lambda,q)$, GA settings $(P,G)$}
\KwOut{policy parameters $\hat\theta$}
Compute LOO neighbour lists of all $x_i$ up to rank $K$\;
Build the vote tensor $V$, the tie-break tensor $T$, and the tables $P,A$\;
Compute and standardise the meta-features $z(x_i)$\;
$k^\star\gets\argmax_k \frac1n\sum_i A_{ik}$\;
Initialise the population with the constant policies for $k^\star$ and a fixed grid of $k$, the rest uniform in $\Theta$\;
\For{$g=1,\dots,G$}{
  Select parents by tournament; apply SBX and annealed Gaussian mutation; reflect into $\Theta$\;
  Evaluate $F(\theta)=\frac1n\sum_i A_{i,k_\theta(x_i)}-\lambda\Omega(\theta)$ for every offspring\;
  Keep the $2$ best parents and the best offspring (elitism)\;
  \lIf{no improvement during $25$ generations}{\textbf{break}}
}
\Return{best genome $\hat\theta$ found}\;
\BlankLine
\textbf{Prediction} for a query $x$: find its $K$ neighbours, compute $z(x)$, set $k=k_{\hat\theta}(x)$, return $\hat y_k(x)$.
\end{algorithm}

\subsection{Computational complexity}\label{sec:complexity}

\begin{proposition}[Cost]\label{prop:cost}
With brute-force neighbour search in $\R^p$, \eank{} training costs
$O(n^2p+nK\log K+nKC+PG\,n(d+q))$ time and $O(nK+nd)$ memory. Prediction
costs $O(np+KC+d)$ per query, the same order as $k$-NN with $k=K$.
\end{proposition}
\begin{proof}
The neighbour search costs $O(n^2p)$, plus $O(nK\log K)$ to sort the $K$ retained
neighbours. Building the cumulative sums $V,T$ and the argmax over classes costs
$O(nKC)$; only $A$ (size $nK$) and $z$ (size $nd$) are kept. A fitness
evaluation computes $n$ scores in $O(nd)$, looks up $n$ entries of $A$, and sums
$nq$ absolute differences. The GA performs at most $P(G+1)$ evaluations.
Prediction needs one neighbour query, the $K$ cumulative votes and one inner
product.
\end{proof}

\section{Theoretical analysis}\label{sec:theory}

All proofs are given in \Cref{app:proofs}.

\subsection{Dominance over global \texorpdfstring{$k$}{k} and stability}

\begin{proposition}[Seeded dominance]\label{prop:dominance}
Assume $(K-1)\sigma(-B_b)<\tfrac12$ and $(K-1)\sigma(B_b)>K-\tfrac32$; this holds
for $B_b=6$ whenever $K\le 202$. Then for every $k\in\{1,\dots,K\}$ there is
$\theta_k\in\Theta$ with $k_{\theta_k}\equiv k$, and the output of seeded \eank{}
satisfies
\[
F(\hat\theta)\;\ge\;\max_{1\le k\le K}\ \frac1n\sum_{i=1}^n A_{ik}
\;=\;\text{LOO accuracy of the best global }k.
\]
\end{proposition}

The guarantee concerns the \emph{training} criterion. Whether the improvement
carries over to unseen data is the question of generalisation, treated below
and measured in \Cref{sec:gap}.

\begin{proposition}[Lipschitz neighbourhoods]\label{prop:lipschitz}
For all $x,x'$ and $\theta=(w,b)$,
\(
|k_\theta(x)-k_\theta(x')|\le \tfrac{K-1}{4}\,\|w\|_2\,\|z(x)-z(x')\|_2+1 .
\)
In particular, on $\Theta$ the change in $k$ is at most
$\tfrac{K-1}{4}B_w\sqrt d\,\|z(x)-z(x')\|_2+1$.
\end{proposition}

Points with similar neighbourhood descriptions therefore receive similar
neighbourhood sizes. The penalty $\Omega$ adds a data-dependent version of this
property on the training graph.

\subsection{Convergence of the genetic search}

$F$ takes finitely many values, because $(k_\theta(x_i))_i$ ranges over a finite
set. However, a maximiser of $F$ may sit on a set of Lebesgue measure zero, for
example a single boundary point. The meaningful target for a stochastic search
on a continuous genome is the \emph{essential supremum}
\[
F^\star_{\mathrm{ess}}=\max\big\{v\in F(\Theta):\ \mathrm{Leb}\{\theta\in\Theta: F(\theta)\ge v\}>0\big\}.
\]

\begin{theorem}[Almost-sure convergence]\label{thm:convergence}
Run the GA of \Cref{alg:eank} without the stagnation stop ($G=\infty$). Let
$F^{\mathrm{best}}_g$ be the best fitness in the population at generation $g$.
Then $F^{\mathrm{best}}_g$ is non-decreasing and there is $\varepsilon>0$, which
depends only on $D$, $\sigma_0$, $\sigma_{\min}$, $\Theta$ and $F$, such that
$\Pb\{F^{\mathrm{best}}_g<F^\star_{\mathrm{ess}}\}\le(1-\varepsilon)^{g}$.
Consequently $F^{\mathrm{best}}_g\ge F^\star_{\mathrm{ess}}$ for all $g$ large
enough, almost surely.
\end{theorem}

\begin{corollary}\label{cor:ess}
$F^\star_{\mathrm{ess}}\ge\max_k\frac1n\sum_iA_{ik}$. Hence even an
\emph{unseeded} \eank{} reaches the best global-$k$ LOO accuracy almost surely.
Seeding achieves this at generation $0$.
\end{corollary}

This is the elitist analogue, for a continuous genome, of the result of
\citet{rudolph1994}: canonical GAs without elitism do not converge to the
optimum, while elitist ones do.

\subsection{Capacity of the policy class and generalisation}

Fix the training sample $\D$. Then $z(\cdot)$ and the map $x\mapsto(\hat y_k(x))_{k\le K}$
are fixed functions, and the adaptive rule $h_\theta=h_{k_\theta}$ is indexed by
$\theta\in\R^{d+1}$ only.

\begin{lemma}[Growth function]\label{lem:growth}
For any $m$ points $u_1,\dots,u_m\in\R^p$, the number of distinct vectors
$\big(k_\theta(u_1),\dots,k_\theta(u_m)\big)$ realised as $\theta$ ranges over
$\R^{d+1}$ is at most
\[
\Pi(m)\;\le\;\sum_{j=0}^{d+2}\binom{m(K-1)}{j}\;\le\;\Big(\frac{e\,m(K-1)}{d+2}\Big)^{d+2}\qquad(m(K-1)\ge d+2).
\]
\end{lemma}

\begin{theorem}[Uniform generalisation of adaptive policies]\label{thm:generalisation}
Let $\mathcal V=\{(u_j,v_j)\}_{j=1}^m$ be an i.i.d.\ sample from $\Pb$,
independent of $\D$, and let $\hat R_{\mathcal V}(\theta)$ be the error of
$h_\theta$ on $\mathcal V$. With probability at least $1-\delta$ over $\mathcal V$,
simultaneously for all $\theta\in\R^{d+1}$,
\[
R(h_\theta)\ \le\ \hat R_{\mathcal V}(\theta)+\sqrt{\frac{32}{m}\Big[(d+2)\ln\frac{e\,m(K-1)}{d+2}+\ln\frac8\delta\Big]} .
\]
\end{theorem}

\begin{remark}[Policy versus per-instance encodings]\label{rem:direct}
The bound holds uniformly in $\theta$. It therefore covers any $\hat\theta$
selected by any procedure, including a GA run on $\mathcal V$ itself, at a price
of order $\sqrt{d\log(mK)/m}$ that does not depend on the number of training
points or on the number of fitness evaluations. A direct encoding
$(k_1,\dots,k_n)\in\{1,\dots,K\}^n$ has no such guarantee: up to $K^n$
assignments can be realised on the training points, so selecting among them on
the same sample can overfit arbitrarily. This motivates the policy design of
\eank{}.
\end{remark}

\begin{remark}[LOO fitness]
\eank{} uses the LOO accuracy on $\D$ rather than an independent $\mathcal V$, in
order to use all the data. \Cref{thm:generalisation} then does not apply
verbatim, because the meta-features and the candidates depend on the same sample.
\Cref{lem:growth} still bounds the effective number of distinct candidates by a
polynomial of degree $d+2$, which limits the optimism of LOO selection. We
quantify this optimism empirically in \Cref{sec:gap}. A validation-based
variant, to which \Cref{thm:generalisation} applies exactly, is a direct
extension.
\end{remark}

\section{Experiments}\label{sec:experiments}

\subsection{Setup}

\paragraph{Datasets.} We use twelve public classification datasets
(\Cref{tab:datasets}). Iris, Wine, Breast cancer and Digits come from
scikit-learn \citep{pedregosa2011}; the other eight come from the UCI repository
\citep{uci}. Constant features are removed, and all features are standardised
using training-fold statistics.

\paragraph{Methods.} All methods share the same neighbour search, tie-breaking
rule and $K=50$, so differences come only from the choice of $k$.
(1)~\emph{kNN ($k{=}5$)}: the common default.
(2)~\emph{kNN-LOO}: global $k$ by exhaustive LOO search, which is the exact
optimum of the global-$k$ criterion and therefore an upper bound for grid,
random, Bayesian or evolutionary search over $k$.
(3)~\emph{wkNN-LOO}: distance-weighted voting \citep{dudani1976} with $k$ chosen
by exhaustive LOO.
(4)~\emph{GA-k}: a classical GA over the single integer $k$ (population 10,
20 generations), representative of
\citep{prasetio2020,mcroberts2016}.
(5)~\emph{LocalK}: a two-stage local-$k$ method in the spirit of
\citet{wettschereck1994,zhang2018ktree}. Each training point gets the
LOO-correct $k$ closest to $k^\star$, and a query uses the median of the values
of its $5$ nearest training points.
(6)~\emph{\eank{}} with $\lambda=0.05$, $q=5$, $m=10$, $M=30$ and the GA settings of
\Cref{sec:method}. These defaults were fixed before the benchmark was run and
were not tuned per dataset.

\paragraph{Protocol.} We use $5\times$ repeated stratified $5$-fold
cross-validation, i.e.\ 25 train/test splits per dataset and 300 in total. All
model selection (choice of $k$, evolution of $\theta$, standardisation) happens
inside the training fold. The test fold is used only for scoring, so the
procedure is nested with respect to every tuned quantity \citep{cawley2010,varma2006}. We
report accuracy, macro-F1, balanced accuracy and wall-clock times. Across
datasets we use the Friedman test with the Nemenyi critical difference, and
Wilcoxon signed-rank tests with Holm correction \citep{demsar2006,derrac2011}. Within a
dataset we use the corrected resampled $t$-test of \citet{nadeau2003}, which
accounts for the overlap between training sets.

\begin{table}[t]
\centering
\caption{Datasets: number of instances $n$, features $p$ (after removing constant ones), classes $C$, and imbalance ratio (largest / smallest class).}\label{tab:datasets}
\small
\begin{tabular}{lrrrr}
\toprule
Dataset & $n$ & $p$ & $C$ & Imb. ratio\\
\midrule
\input{tables/datasets.tex} 
\bottomrule
\end{tabular}
\end{table}

\subsection{Controlled study: spatially heterogeneous neighbourhoods}\label{sec:synthetic}

We first test the hypothesis that motivates \eank{}: an adaptive neighbourhood
helps when the optimal amount of smoothing varies across the input space. The
synthetic distribution has two regions of equal mass. The left region is a
noiseless $4\times4$ checkerboard on $[0,1]^2$, a fine structure that calls for
small $k$. The right region is a linear boundary on $[1.5,3.5]\times[0,1]$ whose
labels are flipped with probability $\eta$, a smooth structure with noisy labels
that calls for large $k$. We train on $n=1200$ points and test on $4000$
independent points, with 10 replications per noise level.

\begin{figure}[t]
\centering
\includegraphics[width=\linewidth]{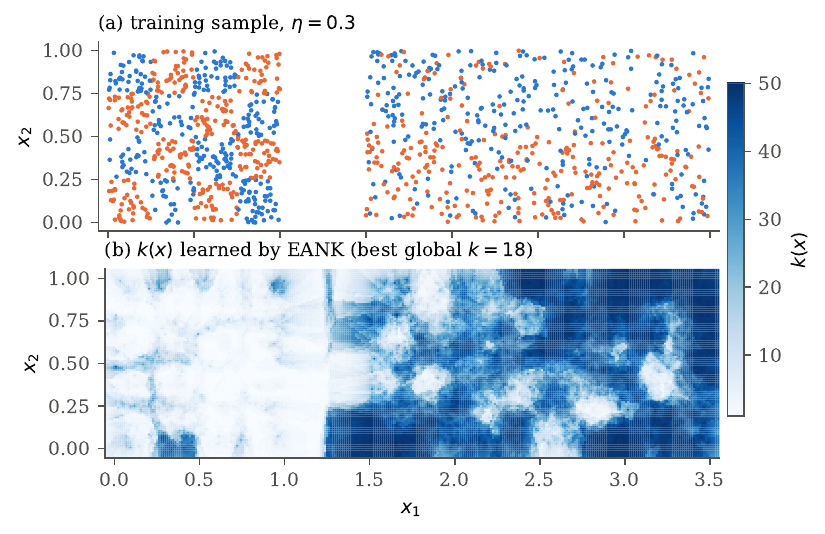}
\caption{Synthetic heterogeneous benchmark ($\eta=0.3$). (a)~Training sample: noiseless checkerboard on the left, linear boundary with 30\% label noise on the right.
(b)~Neighbourhood size $k(x)$ chosen by the learned \eank{} policy over the
input space. The policy selects small neighbourhoods in the fine-grained
checkerboard and large ones in the noisy smooth region. The best single global
value found by exhaustive LOO search is a compromise between the two.}
\label{fig:kmap}
\end{figure}

\begin{table}[t]
\centering
\caption{Synthetic benchmark: test accuracy (\%, mean over 10 replications) as a
function of the label-noise rate $\eta$ in the smooth region. $p$: two-sided
Wilcoxon signed-rank test, \eank{} vs.\ kNN-LOO. Last two columns: global $k$ selected
by kNN-LOO, and median $k$ used by \eank{} in the left / right regions.}\label{tab:synthetic}
\small
\begin{tabular}{ccccccc}
\toprule
$\eta$ & kNN-LOO & LocalK & \eank{} & $p$ & $k$ (global) & $k$ \eank{} (L / R)\\
\midrule
\input{tables/synthetic.tex} 
\bottomrule
\end{tabular}
\end{table}

\begin{figure}[t]
\centering
\includegraphics[width=\linewidth]{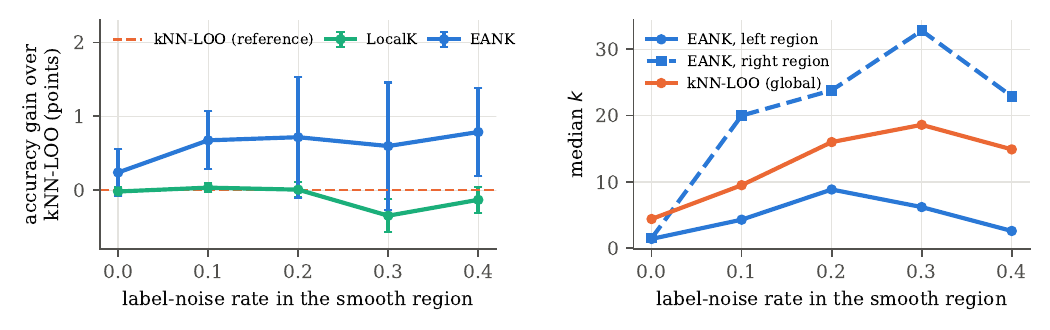}
\caption{Synthetic benchmark. Left: test-accuracy gain over the best global $k$ (kNN-LOO) versus noise rate (mean $\pm$ s.d.\ over 10 paired replications). Right: median neighbourhood size in each region.}
\label{fig:synthetic}
\end{figure}

\Cref{fig:kmap,tab:synthetic,fig:synthetic} confirm the hypothesis. Without
label noise, both the global and the adaptive rules pick small neighbourhoods.
As $\eta$ grows, the global optimum moves to $k\approx15$--$19$, a compromise
that over-smooths the checkerboard. \eank{} instead keeps a small $k$ in the left
region and, for $\eta\ge0.1$, uses a neighbourhood three to nine times larger on
the right.
\eank{} is the most accurate method at every noise level. The gain over the
best global $k$ ranges from $0.2$ to $0.8$ points and is significant at
$\eta\in\{0.1,0.2,0.4\}$. The two-stage LocalK baseline does not improve on the
global rule here: per-point LOO targets are too noisy to transfer.

\subsection{Benchmark results}

\begin{table}[t]
\centering
\caption{Test accuracy (\%, mean\,$\pm$\,s.d.\ over 25 nested splits). Bold: best per row.
An arrow on \eank{} would mark a significant difference with kNN-LOO
(corrected resampled $t$-test, $\alpha=0.05$); none was observed.
Mean rank: average Friedman rank over the 12 datasets (lower is better).}\label{tab:main}
\scriptsize
\setlength{\tabcolsep}{3.5pt}
\begin{tabular}{lcccccc}
\toprule
Dataset & kNN ($k$=5) & kNN-LOO & wkNN-LOO & GA-k & LocalK & \eank{}\\
\midrule
\input{tables/main_acc.tex} 
\bottomrule
\end{tabular}
\end{table}

\Cref{tab:main} reports test accuracy. \eank{} obtains the best average rank
(2.75), followed by GA-k (2.88), wkNN-LOO and LocalK (3.42), kNN-LOO (3.54) and
the default $k=5$ (5.00). The Friedman test rejects equality of the six methods
($\chi^2_F=11.65$, $p=0.040$). The Nemenyi critical difference is $2.18$, so only
the gap between \eank{} and $k=5$ (2.25) is significant.
\Cref{tab:stats} gives the pairwise picture. \eank{} wins on 11 of 12 datasets
against $k=5$ (Holm-adjusted $p=0.005$). Against the tuned baselines it wins on 6
to 8 datasets and loses on 4 to 5, and none of these differences is significant,
neither across datasets nor within any single dataset. The largest gains over
kNN-LOO occur on Ionosphere ($+0.92$ points), Pima ($+0.62$), Seeds ($+0.57$) and
Glass ($+0.47$). The largest losses occur on Haberman ($-1.31$), Iris ($-0.53$) and
Wine ($-0.46$).

\begin{table}[t]
\centering
\caption{Pairwise comparison of \eank{} with each baseline over the 12 datasets
(mean accuracy): wins/ties/losses, Wilcoxon signed-rank $p$-value, Holm-adjusted
$p$-value, and number of datasets on which \eank{} is significantly
better/worse according to the corrected resampled $t$-test.}\label{tab:stats}
\small
\begin{tabular}{lcccc}
\toprule
Baseline & W/T/L & $p$ & $p_{\text{Holm}}$ & sig.\ better/worse\\
\midrule
\input{tables/stats.tex} 
\bottomrule
\end{tabular}
\end{table}

\begin{table}[t]
\centering
\caption{Other metrics, averaged over the 12 datasets (ranks: lower is better), and mean wall-clock time per training fold on one CPU core.}\label{tab:other}
\scriptsize
\setlength{\tabcolsep}{4pt}
\begin{tabular}{lcccccc}
\toprule
 & kNN ($k$=5) & kNN-LOO & wkNN-LOO & GA-k & LocalK & \eank{}\\
\midrule
\input{tables/other_metrics.tex} 
\bottomrule
\end{tabular}
\end{table}

On macro-F1 and balanced accuracy (\Cref{tab:other}), \eank{} ranks second, just
behind distance-weighted kNN-LOO. The fitness optimises plain accuracy, so this
is consistent with the objective rather than evidence of an imbalance-specific
benefit. Training takes about $0.11$\,s per fold, against $0.01$\,s for
exhaustive global search. Prediction is only about $40\%$ slower than $k$-NN,
because the meta-features reuse the neighbour list of the query.

\subsection{In-sample versus out-of-sample gains}\label{sec:gap}

\Cref{prop:dominance} guarantees that \eank{} improves the LOO criterion over the
best global $k$. On average the improvement is $+1.52$ points (up to $+5.05$ on
Glass). The test-set improvement over kNN-LOO, however, averages only $+0.01$
points. The correlation between the two across datasets is $0.37$: datasets
where the policy finds more in-sample structure (Glass, Ionosphere, Sonar) tend
to be those where \eank{} helps most, but most of the LOO gain is selection
optimism. This is the empirical face of the LOO caveat after
\Cref{thm:generalisation}. Even a policy with eight parameters can fit chance
fluctuations of the LOO correctness table on samples of a few hundred points.
The smoothness penalty mitigates this optimism but does not remove it
(\Cref{sec:ablation}).

\begin{figure}[t]
\centering
\includegraphics[width=\linewidth]{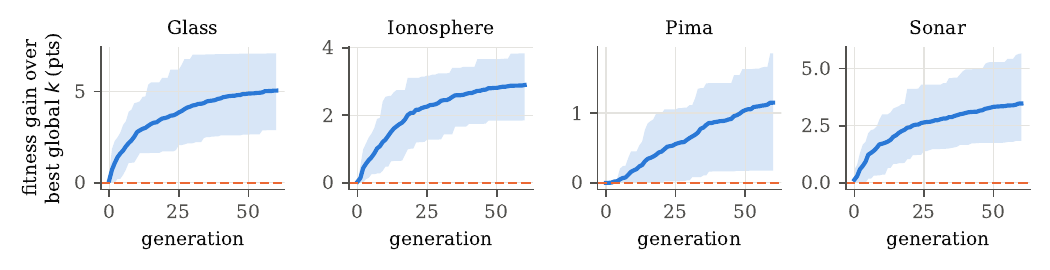}
\caption{GA convergence on four datasets: fitness gain over the best global $k$ (LOO, points), mean and 10--90\% band over 25 folds. The dashed line marks the seeded starting point (\Cref{prop:dominance}).}
\label{fig:convergence}
\end{figure}

\begin{figure}[t]
\centering
\includegraphics[width=\linewidth]{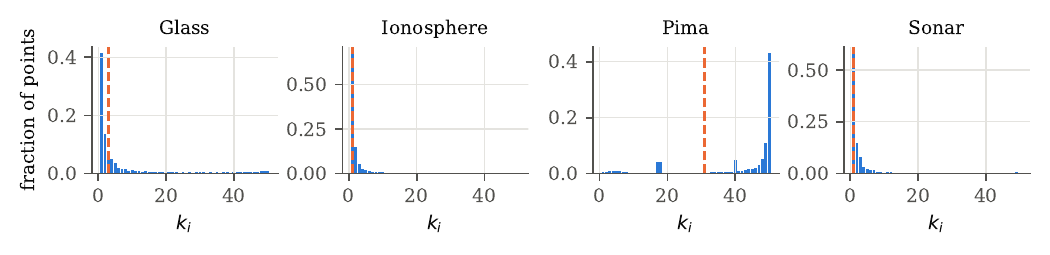}
\caption{Distribution of the neighbourhood sizes $k_i$ assigned by \eank{} to training points, pooled over folds. Dashed: median global $k$ chosen by kNN-LOO.}
\label{fig:kdist}
\end{figure}

\subsection{Ablation study}\label{sec:ablation}

\begin{table}[t]
\centering
\caption{Ablations over the 12 datasets: mean test accuracy (\%), mean rank among
the six variants, mean LOO fitness gain over the best global $k$ (points),
mean within-fold standard deviation of test-time $k$, mean number of fitness
evaluations, and mean training time (ms).}\label{tab:ablation}
\small
\begin{tabular}{lcccccc}
\toprule
Variant & Acc. & Rank & LOO gain & s.d.($k$) & \#eval & Time\\
\midrule
\input{tables/ablation.tex} 
\bottomrule
\end{tabular}
\end{table}

\Cref{tab:ablation} isolates each design choice.
\begin{itemize}
\item \textbf{Smoothness} ($\lambda=0$). Removing $\Omega$ increases the LOO gain
  ($+1.84$ vs.\ $+1.52$) and more than doubles the variability of $k$ ($12.6$
  vs.\ $5.2$), while lowering test accuracy and giving the worst rank. This is the
  overfitting mechanism of \Cref{sec:gap}, and the regulariser counters it.
\item \textbf{Search} (random search with the same budget). Random search finds
  only a quarter of the GA's LOO improvement ($+0.36$). It nevertheless reaches a
  similar test accuracy: the GA's extra in-sample progress is partly optimism.
\item \textbf{Seeding.} Without constant-policy seeds the GA still exceeds the
  global optimum in-sample (\Cref{cor:ess}), but it converges more slowly and
  generalises slightly worse.
\item \textbf{Meta-features.} Label-based features (entropy, margin, agreement)
  carry most of the useful signal. Density features alone are weaker, and the
  full set is best, although the differences between feature sets are small.
\end{itemize}

\subsection{What do the learned policies say?}

\begin{figure}[t]
\centering
\includegraphics[width=\linewidth]{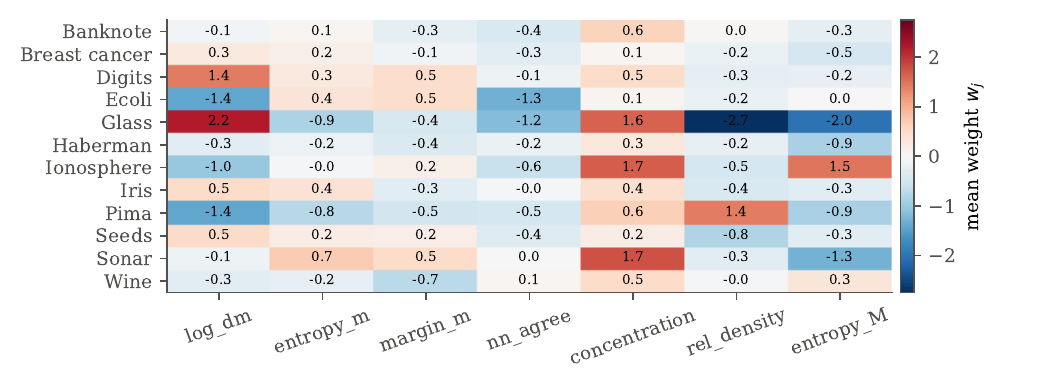}
\caption{Mean policy weights $w_j$ per dataset (25 folds). Positive weights increase $k$.}
\label{fig:weights}
\end{figure}

In 21\% of the folds, the GA returns the seeded constant policy unchanged, i.e.\
it finds no adaptive policy better than the best global $k$. This happens in
60\% of the folds on Banknote, where all methods exceed 99.8\% accuracy, and
never on Glass, Ionosphere or Sonar. \eank{} therefore falls back to the global
rule when there is nothing to adapt to. Among non-constant policies, the
coefficient of \texttt{concentration} is positive in 78\% of the folds: points
with an almost-duplicate nearest neighbour get small neighbourhoods. The
coefficient of \texttt{nn\_agree} is negative in 71\%: when the 1-NN disagrees
with the local majority, a larger vote is used. Both patterns match the
intuition behind adaptive $k$. Other coefficients change sign across datasets
(\Cref{fig:weights}), which reflects dataset-specific trade-offs.

\FloatBarrier
\section{Discussion and limitations}\label{sec:discussion}

\paragraph{When does adaptive $k$ pay off?} The synthetic study shows a clear
and consistent benefit when the optimal smoothing genuinely varies across the
input space. On the twelve standard benchmarks, which are small and mostly
homogeneous, \eank{} is at least as good as tuned global $k$ on average and has
the best mean rank. Its advantage is not statistically significant, however, and
it loses on some datasets (Haberman, Iris, Wine). We regard this as an honest
picture: much of the benefit reported for local-$k$ methods may depend on the
evaluation protocol. Our protocol is fully nested, uses corrected tests, and
gives the baselines the same tie-breaking rule and the exact global optimum.

\paragraph{Limitations.} (i)~The fitness is the LOO accuracy on the training
set, which is optimistic (\Cref{sec:gap}). Our generalisation guarantee is exact
only for a held-out fitness. (ii)~The policy is a monotone function of a linear
score. Non-monotone dependencies, such as small $k$ at both very high and very
low density, require a richer policy, for instance a small network, which
increases $d$ in \Cref{thm:generalisation}. (iii)~The datasets are small to
medium ($n\le1797$); with brute-force search the $O(n^2p)$ neighbour
computation dominates for large $n$, and approximate search would be needed.
(iv)~We optimise accuracy; imbalance-aware or calibration-aware fitness
functions are natural but untested here.

\paragraph{Future work.} Promising directions are: held-out or
cross-fitted fitness to close the in-sample/out-of-sample gap; multi-objective
evolution that trades accuracy against smoothness and robustness, giving
Pareto fronts of policies; class-aware meta-features for imbalanced data; and
tracking distribution shift by re-evolving the policy online, warm-started from
the previous population.

\section{Conclusion}

We argued that using a genetic algorithm to search for a single global $k$
cannot beat exhaustive search, and we proposed to evolve a neighbourhood
\emph{policy} instead. \eank{} learns an interpretable, low-dimensional map
from local neighbourhood statistics to an instance-dependent $k$. A
pre-computed vote table makes each fitness evaluation cost $O(n)$. We showed
that the method dominates the best global $k$ on its training criterion, that
the GA converges almost surely, and that its policy class admits a
generalisation bound free of the sample size. Experiments show clear gains when
the optimal neighbourhood is spatially heterogeneous, and performance on par
with the best tuned baselines, with the best mean rank, on twelve standard
datasets. They also identify the gap between in-sample and out-of-sample
improvement as the main obstacle that future adaptive-neighbourhood methods must
address.

\paragraph{Code and data availability.} The Python implementation of \eank{},
the baselines and the scripts that reproduce every table and figure of this paper
are available from the corresponding author upon request. All datasets used in
this study are publicly available (\Cref{tab:datasets}).

\FloatBarrier
\bibliographystyle{plainnat}
\bibliography{references}

\appendix
\section{Proofs}\label{app:proofs}

\begin{proof}[Proof of \Cref{prop:dominance}]
For $k\in\{2,\dots,K-1\}$, take $w=0$ and $b=\operatorname{logit}\big((k-1)/(K-1)\big)$.
Then $(K-1)\sigma(b)=k-1$ exactly and $k_\theta\equiv k$, with
$|b|\le\operatorname{logit}\big((K-2)/(K-1)\big)=\ln(K-2)<B_b$ for $K\le202$.
For $k=1$ take $b=-B_b$: since $(K-1)\sigma(-B_b)<\tfrac12$, the rounding gives
$0$ and $k_\theta\equiv1$. For $k=K$ take $b=B_b$: since $(K-1)\sigma(B_b)>K-\tfrac32$,
the rounding gives $K-1$. Constant policies have $\Omega=0$, so
$F(\theta_k)=\frac1n\sum_iA_{ik}$. The initial population contains
$\theta_{k^\star}$, and elitist replacement never removes the current best
genome, so the best fitness never decreases. For $B_b=6$,
$\sigma(-6)\approx0.00247$, and both conditions hold when $K-1<202.2$.
\end{proof}

\begin{proof}[Proof of \Cref{prop:lipschitz}]
The logistic function is $\tfrac14$-Lipschitz. Hence
$|(K-1)\sigma(s)-(K-1)\sigma(s')|\le\tfrac{K-1}{4}|w^\top(z(x)-z(x'))|$, and
Cauchy--Schwarz bounds the inner product by $\|w\|_2\|z(x)-z(x')\|_2$. Finally,
$|\lfloor a+\tfrac12\rfloor-\lfloor a'+\tfrac12\rfloor|\le|a-a'|+1$ for all reals
$a,a'$. On $\Theta$, $\|w\|_2\le B_w\sqrt d$.
\end{proof}

\begin{proof}[Proof of \Cref{thm:convergence}]
\emph{Monotonicity.} The two best parents enter the merged pool that is
truncated to the $P$ best, so the best fitness never decreases.

\emph{Hitting probability.} Let $A=\{\theta\in\Theta:F(\theta)\ge F^\star_{\mathrm{ess}}\}$,
so $\mathrm{Leb}(A)>0$ by definition. Consider one offspring, whatever its
parents and crossover result $\tilde\theta\in\Theta$. With probability at least
$D^{-(D-1)}$ all $D$ genes are mutated: one gene is forced, and each of the
other $D-1$ is mutated independently with probability $1/D$. Conditional on
this event, $\theta'=\mathcal R(\tilde\theta+\sigma_g S\xi)$, where
$\xi\sim\mathcal N(0,I_D)$, $S=\mathrm{diag}(s_1,\dots,s_D)$ holds the box
widths, and $\mathcal R$ is the coordinate-wise reflection into $\Theta$.
Reflection maps each point $t\in\Theta$ back from at least the pre-image
$t$ itself, so the density of $\theta'$ at $t$ is at least the Gaussian density
of $\tilde\theta+\sigma_gS\xi$ at $t$. Since $|t_j-\tilde\theta_j|\le s_j$,
\[
f_{\theta'}(t)\ \ge\ \prod_{j=1}^D\frac{1}{\sqrt{2\pi}\,\sigma_g s_j}\exp\!\Big(-\frac{1}{2\sigma_g^2}\Big)
\ \ge\ \prod_{j=1}^D\frac{1}{\sqrt{2\pi}\,\sigma_0 s_j}\exp\!\Big(-\frac{1}{2\sigma_{\min}^2}\Big)=:c>0,
\]
using $\sigma_{\min}\le\sigma_g\le\sigma_0$. Therefore every offspring falls in $A$
with probability at least $\varepsilon:=D^{-(D-1)}c\,\mathrm{Leb}(A)>0$,
independently of the past given the parents. Without hitting $A$ the best
fitness stays below $F^\star_{\mathrm{ess}}$, and once $A$ is hit elitism keeps
the best fitness at or above $F^\star_{\mathrm{ess}}$. Hence
$\Pb\{F^{\mathrm{best}}_g<F^\star_{\mathrm{ess}}\}\le(1-\varepsilon)^{Pg}\le(1-\varepsilon)^g$.

\emph{Almost-sure statement.} Since $\sum_g(1-\varepsilon)^g<\infty$, the
Borel--Cantelli lemma shows that $F^{\mathrm{best}}_g\ge F^\star_{\mathrm{ess}}$
eventually, almost surely. (The best fitness may exceed $F^\star_{\mathrm{ess}}$
if the search lands on a Lebesgue-null set of higher fitness, for example
through seeds with $w=0$ and partial mutations that keep some genes unchanged;
the theorem does not exclude this.)
\end{proof}

\begin{proof}[Proof of \Cref{cor:ess}]
Let $\theta_{k^\star}=(0,b^\star)$ with $(K-1)\sigma(b^\star)=k^\star-1$ an
integer (the cases $k^\star\in\{1,K\}$ are treated as in
\Cref{prop:dominance}, with $b^\star$ in the interior of the half-line where the
rounding is constant). The training meta-features are bounded, say
$\|z(x_i)\|\le Z$. By continuity there is an open neighbourhood $U$ of
$\theta_{k^\star}$ on which $|(K-1)\sigma(w^\top z(x_i)+b)-(k^\star-1)|<\tfrac12$
for all $i$. On $U$, $k_\theta\equiv k^\star$, $\Omega=0$, and
$F=\frac1n\sum_iA_{ik^\star}$. Since $\mathrm{Leb}(U\cap\Theta)>0$, the
definition of $F^\star_{\mathrm{ess}}$ gives the claim, and \Cref{thm:convergence}
applies to the unseeded GA.
\end{proof}

\begin{proof}[Proof of \Cref{lem:growth}]
Let $t_j=\operatorname{logit}\big((j-\tfrac12)/(K-1)\big)$ for $j=1,\dots,K-1$.
By \eqref{eq:policy}, $k_\theta(u)=1+\#\{j: w^\top z(u)+b\ge t_j\}$. The vector
$(k_\theta(u_i))_{i\le m}$ is therefore determined by the $m(K-1)$ bits
$\ind[\langle(w,b,1),(z(u_i),1,-t_j)\rangle\ge0]$. These bits are the values of
the homogeneous half-space indicator with normal $(w,b,1)\in\R^{d+2}$ at the
$m(K-1)$ points $(z(u_i),1,-t_j)\in\R^{d+2}$. Homogeneous half-spaces in
$\R^{d+2}$ have VC dimension $d+2$, so by the Sauer--Shelah lemma
\citep{sauer1972} they realise at most $\sum_{j=0}^{d+2}\binom{m(K-1)}{j}$ bit
patterns on these points. The last inequality is the standard bound
$\sum_{j\le h}\binom{N}{j}\le(eN/h)^h$ for $N\ge h$.
\end{proof}

\begin{proof}[Proof of \Cref{thm:generalisation}]
Conditionally on $\D$, consider the error indicators
$\ell_\theta(u,v)=\ind[\hat y_{k_\theta(u)}(u)\neq v]$, $\theta\in\R^{d+1}$. On any
$m$ points, the vector of these indicators is a function of
$(k_\theta(u_i))_{i\le m}$, so the class realises at most $\Pi(m)$ patterns
(\Cref{lem:growth}). The Vapnik--Chervonenkis inequality
\citep{vapnik1971}, in the form of \citet[Thm.~12.5]{devroye1996}, gives
\[\Pb\Big\{\sup_\theta\big|\hat R_{\mathcal V}(\theta)-R(h_\theta)\big|>\epsilon\Big\}\le8\,\Pi(m)\,e^{-m\epsilon^2/32}.\]
Setting the right-hand side to $\delta$ and using \Cref{lem:growth} yields the
bound. The bound holds conditionally on every $\D$, hence also
unconditionally.
\end{proof}

\section{Hyper-parameters and implementation details}\label{app:hyper}

\begin{table}[ht]
\centering
\caption{Default hyper-parameters of \eank{} (identical for all datasets).}
\small
\begin{tabular}{llll}
\toprule
Symbol & Value & Symbol & Value\\
\midrule
$K$ (max.\ neighbours) & $50$ & $P$ (population) & $40$\\
$m$, $M$ (feature scales) & $10$, $30$ & $G$ (generations) & $60$ (patience $25$)\\
$\lambda$, $q$ (smoothness) & $0.05$, $5$ & tournament size & $3$\\
$B_w$, $B_b$ (box) & $4$, $6$ & SBX $p_c$, $\eta$ & $0.9$, $15$\\
elites & $2$ & mutation $\sigma_0\to\sigma_{\min}$ & $0.5\to0.02$ (annealed)\\
\bottomrule
\end{tabular}
\end{table}

For every training fold, $K$ is capped at $n_{\text{train}}-1$, and $m$ and $M$ at
$K$. The implementation uses NumPy and scikit-learn \citep{pedregosa2011}.
Neighbour lists are computed with \texttt{NearestNeighbors}, and the query point
is removed explicitly from its own list, which handles duplicated points. The
random seed of each split is $1000\cdot\text{repetition}+\text{fold}$. The code
rounds with \texttt{numpy.rint} (half to even), which differs from
$\lfloor a+\frac12\rfloor$ only on a Lebesgue-null set of parameters; the
counting argument of \Cref{lem:growth} holds for either convention. The GA
needs on average $2092$ fitness evaluations ($52$ generations) per fold.

\end{document}

%% file: tables/datasets.tex
Iris & 150 & 4 & 3 & 1.0 \\
Wine & 178 & 13 & 3 & 1.5 \\
Breast cancer & 569 & 30 & 2 & 1.7 \\
Digits & 1797 & 61 & 10 & 1.1 \\
Sonar & 208 & 60 & 2 & 1.1 \\
Ionosphere & 351 & 33 & 2 & 1.8 \\
Pima & 768 & 8 & 2 & 1.9 \\
Glass & 214 & 9 & 6 & 8.4 \\
Seeds & 210 & 7 & 3 & 1.0 \\
Banknote & 1372 & 4 & 2 & 1.2 \\
Haberman & 306 & 3 & 2 & 2.8 \\
Ecoli & 336 & 7 & 8 & 71.5 \\

%% file: tables/synthetic.tex
0.0 & 95.18 & 95.17 & \textbf{95.42} & 0.055 & 4.4 & 1.4 / 1.5 \\
0.1 & 89.81 & 89.85 & \textbf{90.49} & 0.004 & 9.5 & 4.3 / 20.0 \\
0.2 & 84.04 & 84.05 & \textbf{84.76} & 0.016 & 16.0 & 8.8 / 23.8 \\
0.3 & 78.73 & 78.39 & \textbf{79.33} & 0.084 & 18.6 & 6.2 / 32.8 \\
0.4 & 73.05 & 72.92 & \textbf{73.84} & 0.010 & 14.9 & 2.6 / 22.9 \\

%% file: tables/main_acc.tex
Iris & 95.20\,$\pm$\,4.2 & \textbf{96.27\,$\pm$\,3.5} & 95.87\,$\pm$\,3.9 & 95.73\,$\pm$\,3.8 & \textbf{96.27\,$\pm$\,3.5} & 95.73\,$\pm$\,3.8 \\
Wine & 95.75\,$\pm$\,3.6 & 96.29\,$\pm$\,3.0 & \textbf{96.86\,$\pm$\,2.7} & 96.52\,$\pm$\,2.5 & 96.29\,$\pm$\,3.0 & 95.83\,$\pm$\,2.6 \\
Breast cancer & 96.59\,$\pm$\,1.7 & 96.63\,$\pm$\,1.5 & 96.49\,$\pm$\,1.5 & \textbf{96.70\,$\pm$\,1.5} & 96.63\,$\pm$\,1.5 & 96.63\,$\pm$\,1.4 \\
Digits & 97.66\,$\pm$\,0.5 & 97.58\,$\pm$\,0.5 & 97.64\,$\pm$\,0.5 & 97.58\,$\pm$\,0.5 & 97.58\,$\pm$\,0.5 & \textbf{97.70\,$\pm$\,0.6} \\
Sonar & 80.96\,$\pm$\,5.8 & 84.89\,$\pm$\,5.1 & 84.79\,$\pm$\,5.2 & 84.89\,$\pm$\,5.1 & 84.79\,$\pm$\,5.1 & \textbf{84.89\,$\pm$\,4.0} \\
Ionosphere & 84.67\,$\pm$\,4.1 & 85.69\,$\pm$\,3.7 & 85.69\,$\pm$\,3.7 & 85.69\,$\pm$\,3.7 & 85.69\,$\pm$\,3.7 & \textbf{86.61\,$\pm$\,4.0} \\
Pima & 73.28\,$\pm$\,3.0 & 75.10\,$\pm$\,2.9 & 75.18\,$\pm$\,2.8 & 75.21\,$\pm$\,3.0 & 75.68\,$\pm$\,3.2 & \textbf{75.73\,$\pm$\,2.7} \\
Glass & 67.47\,$\pm$\,5.4 & 69.82\,$\pm$\,5.6 & 69.36\,$\pm$\,5.6 & 69.63\,$\pm$\,5.7 & \textbf{70.48\,$\pm$\,5.7} & 70.29\,$\pm$\,5.5 \\
Seeds & 92.57\,$\pm$\,3.2 & 92.10\,$\pm$\,4.4 & \textbf{92.67\,$\pm$\,3.6} & 92.00\,$\pm$\,4.1 & 92.10\,$\pm$\,4.4 & \textbf{92.67\,$\pm$\,4.2} \\
Banknote & 99.84\,$\pm$\,0.2 & 99.80\,$\pm$\,0.3 & 99.82\,$\pm$\,0.3 & \textbf{99.85\,$\pm$\,0.2} & 99.80\,$\pm$\,0.3 & 99.81\,$\pm$\,0.3 \\
Haberman & 70.60\,$\pm$\,4.2 & 74.85\,$\pm$\,3.5 & 74.25\,$\pm$\,3.8 & \textbf{75.11\,$\pm$\,3.1} & 75.04\,$\pm$\,3.9 & 73.54\,$\pm$\,3.6 \\
Ecoli & 85.06\,$\pm$\,3.0 & 86.31\,$\pm$\,3.6 & 86.31\,$\pm$\,3.4 & \textbf{86.37\,$\pm$\,3.5} & 86.19\,$\pm$\,3.9 & 86.01\,$\pm$\,3.8 \\
\midrule
Mean acc. & 86.64 & 87.94 & 87.91 & 87.94 & 88.04 & 87.95 \\
Mean rank & 5.00 & 3.54 & 3.42 & 2.88 & 3.42 & \textbf{2.75} \\

%% file: tables/stats.tex
kNN (k=5) & 11/0/1 & 0.001 & 0.005 & 0/0 \\
kNN-LOO & 8/0/4 & 0.569 & 1.000 & 0/0 \\
wkNN-LOO & 6/1/5 & 0.820 & 1.000 & 0/0 \\
GA-k & 6/1/5 & 0.820 & 1.000 & 0/0 \\
LocalK & 7/0/5 & 0.850 & 1.000 & 0/0 \\

%% file: tables/other_metrics.tex
Macro-F1 (mean) & 82.94 & 84.25 & 84.44 & 84.24 & 84.45 & 84.14 \\
Macro-F1 (rank) & 4.92 & 3.71 & \textbf{2.88} & 3.08 & 3.50 & 2.92 \\
Balanced acc. (mean) & 83.05 & 84.30 & 84.38 & 84.28 & 84.47 & 84.08 \\
Balanced acc. (rank) & 5.00 & 3.67 & \textbf{2.92} & 2.96 & 3.38 & 3.08 \\
Fit time (ms) & 3.2 & 9.1 & 8.1 & 10.2 & 7.8 & 113.8 \\
Predict time (ms) & 1.4 & 2.6 & 2.8 & 2.6 & 2.8 & 3.6 \\

%% file: tables/ablation.tex
EANK & 87.95 & 2.25 & +1.52 & 5.24 & 2092 & 114 \\
EANK ($\lambda$=0) & 87.64 & 4.46 & +1.84 & 12.56 & 1829 & 68 \\
EANK (random search) & 87.81 & 3.42 & +0.36 & 4.21 & 1297 & 59 \\
EANK (no seeding) & 87.68 & 3.79 & +1.53 & 7.18 & 2410 & 131 \\
EANK (density feats) & 87.71 & 3.88 & +1.14 & 4.44 & 2051 & 111 \\
EANK (label feats) & 87.91 & 3.21 & +0.98 & 3.02 & 1901 & 101 \\